\documentclass{article}

\usepackage[final, nonatbib]{neurips_2021}

\usepackage[utf8]{inputenc} 
\usepackage[T1]{fontenc}    
\usepackage{hyperref}       
\usepackage{nameref}       
\usepackage{url}            
\usepackage{booktabs}       
\usepackage{amsfonts}       
\usepackage{nicefrac}       
\usepackage{microtype}      
\usepackage{graphicx}
\usepackage{caption}
\usepackage{multirow}
\usepackage{amsmath}
\usepackage{cleveref}%
\usepackage[table]{xcolor}

\newcommand*{\theadc}[1]{\multicolumn{1}{c}{\begin{tabular}{@{}c@{}}#1\end{tabular}}}

\DeclareMathOperator\arctanh{arctanh}
\newcommand{\K}{\mathcal{K}}

\newcommand{\R}{\mathbb{R}}

\definecolor{purplerstg}{HTML}{9F00FF}
\definecolor{greenrstg}{rgb}{0.1607,0.6039,0.4118}
\definecolor{blackgreenrstg}{rgb}{0.0607,0.5039,0.3518}

\definecolor{blind1_blue}{HTML}{005AB5}
\definecolor{blind1_red}{HTML}{DC3220}

\definecolor{blind2_blue}{HTML}{1A85FF}
\definecolor{blind2_violet}{HTML}{D41159}

\crefname{equation}{equation}{equations}
\crefrangelabelformat{equation}{(#3#1#4--#5#2#6)}
\crefmultiformat{equation}{equations (#2#1#3}{, #2#1#3)}{#2#1#3}{#2#1#3}

\title{Discovering Dynamic Salient Regions for Spatio-Temporal Graph Neural Networks}

\makeatletter
\newcommand{\printfnsymbol}[1]{%
  \textsuperscript{\@fnsymbol{#1}}%
}
\author{%
    Iulia 
  Duta\thanks{Equal contribution.} \\
  Bitdefender, Romania\\
  \texttt{{id366}@cam.ac.uk} \\
  \And
  \And
  Andrei Nicolicioiu\printfnsymbol{1}
  \\
  Bitdefender, Romania\\
  \texttt{{anicolicioiu}@bitdefender.com} \\
   \And
   Marius Leordeanu \\
   Bitdefender, Romania \\
\fontsize{6pt}{7pt}Institute of Mathematics of the Romanian Academy \\
   University "Politehnica" of Bucharest \\
   \texttt{marius.leordeanu@imar.ro} \\
}

\begin{document}
\maketitle

\begin{abstract}

Graph Neural Networks are perfectly suited to capture latent interactions between various entities  in the spatio-temporal domain (e.g. videos). However, when an explicit structure is not available, it is not obvious what atomic elements should be represented as nodes. Current works generally use pre-trained object detectors or fixed, predefined regions to extract graph nodes. 
Improving upon this, our proposed model learns nodes that dynamically attach to well-delimited salient regions, which are relevant for a higher-level task, without using any object-level supervision.
Constructing these localized, adaptive nodes gives our model inductive bias towards object-centric representations and we show that it discovers regions that are well correlated with objects in the video. In extensive ablation studies and experiments on two challenging datasets, we show superior performance to previous graph neural networks models for video classification.\\
Code is available at \url{https://github.com/bit-ml/DyReg-GNN}
\end{abstract}



\section{Introduction}

Spatio-temporal data, and videos, in particular, are characterised by an abundance of events that require complex reasoning to be understood. 
In such data, entities or classes exist at multiple scales and in different contexts in space and time, starting from lower-level physical objects, which are well localized in space and moving towards higher-level concepts which define complex interactions. 
%
We need a representation that captures such spatio-temporal interactions at different level of granularity, depending on the current scene and the requirements of the task.
Classical convolutional nets address spatio-temporal processing in a simple and rigid manner, determined only by fixed local receptive fields~\cite{NIPS2016_6203_effective_receptive_field}. Alternatively, space-time graph neural nets~\cite{yan2018spatial_A, rstg} offer a more powerful and flexible approach
modeling complex short and long-range interactions between visual entities.
%
%
%

In this paper, we propose a novel method to enhance vision Graph Neural Networks (GNNs) by an additional capability, missing from any other previous works. That is, to have nodes 
that are constructed for 
spatial reasoning and can adapt to the current input. Prior works are limited to having either nodes attached to semantic attention maps~\cite{glore} or attached to fixed locations such as grids~\cite{NIPS2017_7082_ralational_cuburi, rstg, dosovitskiy2021an_vit}.
Moreover, unlike works that require external object detectors~\cite{wang2018videos_gupta2}
our method relies on a learnable mechanism to adapt to the current input.

We propose a method that learns to discover salient regions, well-delimited in space and time, that are useful for modeling interactions between various entities. Such entities could be single objects, parts or groups of objects that perform together a simple action. Each node learns to associate by itself to such salient regions, thus the message passing between nodes is able to model object interactions more effectively. 
For humans, representing objects is a core knowledge system \cite{spelke2000core} and to emphasize them in our model, we predict salient regions~\cite{2010_objects} that give a strong inductive bias towards modeling them.\looseness=-1

Our method, Dynamic Salient Regions Graph Neural Network (\textbf{DyReg-GNN}) improves the relational processing of videos by learning to discover salient regions that are relevant for the current scene and task. Note that the model learns to predict regions only from the weak supervision given by the high-level video classification loss, without supervision at the region level. Our experiments convincingly show that the regions discovered are well correlated with the objects present in the video, confirming the intuition that action recognition should be strongly related to salient region discovery. The capacity to discover such regions makes DyReg-GNN an excellent candidate model for tackling tasks requiring spatio-temporal reasoning.




\noindent \textbf{Our main contributions} are summarised as follow:
\begin{enumerate}
\setlength\itemsep{-0.1em}
\item We propose a novel method to \textbf{augment spatio-temporal GNNs} by an additional capability: that of learning to create localized nodes suited for spatial reasoning, that adapt to the input.

\item The salient regions discovery \textbf{enhance the relational processing} for high-level video classification tasks:
creating GNN nodes from predicted regions obtains superior performance compared to both using pre-trained object detectors or fixed regions
\item Our model leads to \textbf{unsupervised salient regions discovery}, a novelty in the realm of GNNs: it predicts such regions in videos, with only weak supervision at the video class level. We show that regions discovered are well correlated with actual physical object instances.


\end{enumerate}

\section{Related work}
\paragraph{Graph Neural Networks in Vision.} 
GNNs have been recently used in many domains where the data has a non-uniform structure ~\cite{DBLP:journals/corr/BrunaZSL13_bruna_lecun_spectral, battaglia2016interaction, pmlr-v70-gilmer17a, li2018learning}. 
In vision tasks, it is important to model the relations between different entities appearing in the scene \cite{ Baradel_2018_ECCV_orn, qi2018learning} and GNNs have strong inductive biases towards relations~\cite{battaglia2018relational, Jegelka_reason_align}, thus they are perfectly suited for modeling interactions between visual instances. Since an explicit  structure  is  not  available in the video, it is of critical importance to establish what atomic elements should be represented as graph nodes. As our main contribution revolves around the creation of nodes, we analyse other recent GNN methods regarding the type of information that each node represents, and group them into two categories, \textit{semantic} and \textit{spatial}.

The approaches of \cite{glore, li2018beyond_segm_anch,sgr_symbolic_graph, latentgnn, Kipf2020Contrastive, kipf_slots} capture the purely \textit{semantic} interactions by reasoning over global graph nodes, each one receiving information from all the points in the input, regardless of spatio-temporal position. In \cite{glore} the nodes assignments are predicted from the input, while in \cite{li2018beyond_segm_anch} the associations between input and nodes are made by a soft clusterization. The work of \cite{kipf_slots}
discovers different representation groups by using an iterative clusterization based on self-attention similarity. 

The downside of these semantic approaches is that individual instances, especially those belonging to the same category, are not distinguished in the graph processing. This information is essential in tasks such as capturing human-object interactions, instance segmentation or tracking. 

Alternatively, multiple methods, including ours, favour modeling instance interactions by defining \textit{spatial} nodes associated with certain locations. We distinguish between them by how they extract the nodes from spatial location: as fixed regions or points
\cite{chen20182_a2nets, graph_tracking}, 
or detected object boxes~\cite{herzig2019spatio, zhang2019structured, dynamic_objects_bmvc, CVPR2020_SomethingElse, differentiable_scene_graph}. 
The method \cite{NIPS2017_7082_ralational_cuburi} creates nodes from every point in 2D convolutional features maps, while Non-Local~\cite{wang2018non_local} uses self-attention~\cite{vaswani2017attention} between all spatio-temporal positions to capture distant interactions. Further, ~\cite{rstg} extract nodes from larger fixed regions at different scales and processes them recurrently. 
Recent methods based on Transformer~\cite{ramachandran2019stand, dosovitskiy2021an_vit, carion2020end_detr} also model the interactions between fixed locations on a grid using self-attention.
In \cite{wang2018videos_gupta2}, nodes are created from object boxes extracted by an external detector and are processed using two different graph structures, one given by location and one given by nodes similarity. A related approach is used in \cite{dynamic_objects_bmvc} in a streaming setting while \cite{hopper} learns to hop over unnecessary frames. 
Hybrid approaches use nodes corresponding to points and object features \cite{sun2018actor_schmid, girdhar2019video_transformer_carreira_az} or propagate over both semantic and spatial nodes~\cite{visual_reasoning_feifei, simbolic_graf_eccv, fu2019_dual_att_segm}.

However, methods that rely on external modules trained on additional data, such as object detectors, 
are too dependent on the module's performance.
They are unable to adapt to the 
current problem, being limited to the set of pre-defined annotations designed for another task.
Differently, our module is optimized to discover regions useful for the current task, using only the video classification signal.

Recently, the method \cite{rahaman2020s2_rms_bengio} uses multiple position-aware nodes that take into account the spatial structure. This makes it more suitable for capturing instances, but the nodes have associated a static learned location, where each one is biased towards a specific position regardless of the input.
On the other hand, we dynamically assign a location for each node, based on the input, making the method more flexible to adapt to new scenes.

\begin{figure*}[t]
\centering
\includegraphics[width=\textwidth]{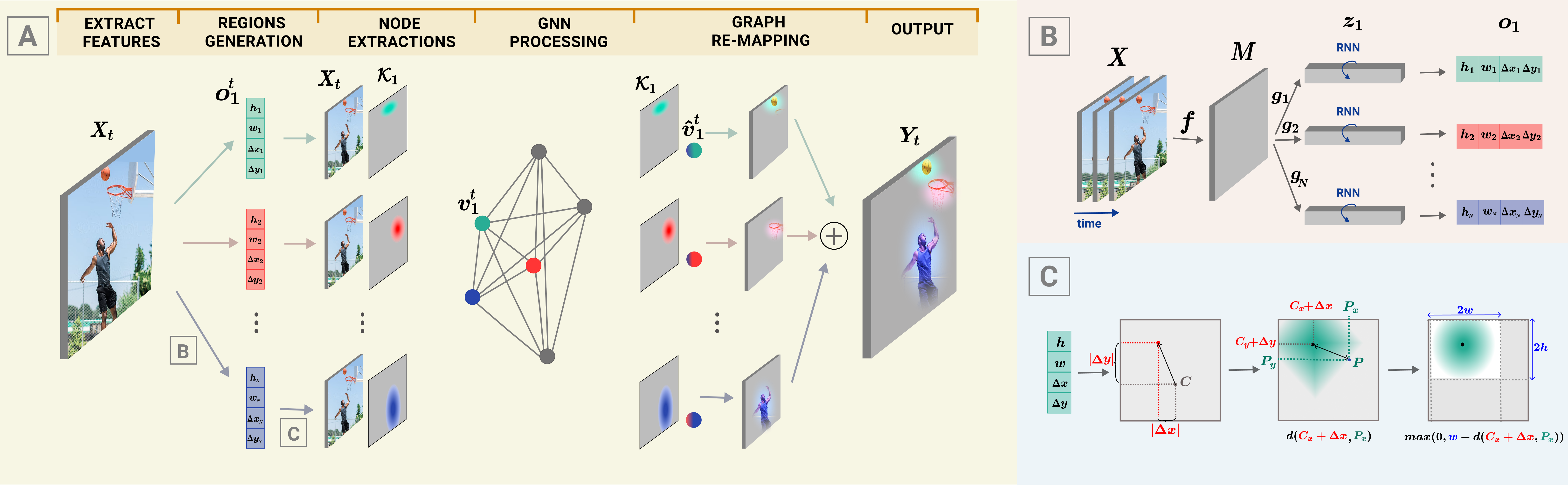}
\caption{ (\textbf{Left}) DyReg-GNN extracts localized node useful for relational processing of videos. For each node $i$, from the features $X_t$, we predict params $\mathbf{o}_i$ denoting the location and size of a region. They define a kernel $K_i$, used to extract  the localized features $\mathbf{v}_i$ from the corresponding region of $X_t$. We process the nodes with a spatio-temporal GNN and project each node $\mathbf{\hat{v}}_i$ into its initial location. 
\textbf{(Right)} B) Node Region Generation: 
Functions $f$ and $\{g_i\}$ generate the regions params $\mathbf{o}_i$; $f$ extracts a latent representation shared between nodes, while each $g_i$ has different params for each node $i$. 
C) Node Features Extraction: 
Each $\mathbf{o}_i$ creates a kernel that is used in a differentiable pooling w.r.t. $\mathbf{o}_i$. This allows us to optimize the generation of these regions' params from the final classification loss, resulting in an unsupervised discovery of salient regions.
}
\label{fig:main_arch}
\end{figure*}

\paragraph{Dynamic Networks.}

Several works use second-order computations by dynamically predicting different parts of their model from the input, instead of directly optimising parameters.
Our work is related to STN~\cite{spatial_transformer} that aggregates features by
interpolating from an area given by a predicted global transformation and to the differentiable pooling used in some object detectors~\cite{Dai2016ROI_warp, he2017mask_rcnn, precise_roi_pool2018}.
The method \cite{jia2016dynamic_filters}, replaces the 
parameters in a standard convolution with weights predicted from the input, resulting in a dynamically generated filter. Deformable convolutions \cite{dai2017deformable, zhu2019deformablev2} predict, based on the input, an offset for each position in the convolutional kernel. Similar, \cite{zhang2020dynamic_torr} use the same idea of predicting offsets but in a graph formulation.
The common topic of these methods is to predict dynamically a support for all points in a convolutional operation while we dynamically generate the input for a set of nodes designed to process high-level interactions. Related ideas, involving high-level processing of a small set of powerful modules, is also highlighted in \cite{goyal2019recurrent_rim_bengio} and \cite{rahaman2020s2_rms_bengio}.

\paragraph{Unsupervised Object Representations.}

There is an entire area of work devoted to extracting representations centered on objects~\cite{greff2020binding} in a fully unsupervised setting \cite{multi-object, burgess2019monet, space, kipf_slots}. They are successful in leveraging a reconstruction task to decompose the scene into objects, for synthetic images. In \cite{better_slots} it is shown that representations learned from unsupervised decomposition are also helpful in relational reasoning tasks.
Methods for generating unsupervised keypoints or entities \cite{unsup_obj_cond_img, unsup_obj_keypoints_control, causal_disc_videos_neurips2020, entity_abstraction_RL} have been generally used in synthetic setting.  The method \cite{unsup_obj_cond_img} generates keypoints from real images of people and faces 
but they use an image reconstruction objective that could not be aligned with the downsteam task.
Our goal is to relate spatio-temporal entities, but without enforcing a clear decomposition of the scene into objects. This allows us to use a simpler but effective method that learns from classification supervision of real-world videos and obtain representations that are correlated to objects.

\paragraph{Activity Recognition.}
Video classification has been influenced by methods designed for 2D images \cite{yue2015beyond, donahue2015long, ma2018ts,zhou2018temporal_trn_torralba}. More powerful 3D convolutional networks 
have been later proposed \cite{carreira2017quo}, while other methods factorise the 3D convolutions \cite{xie2018rethinking, tran2018closer, tran2019video} bringing both computational speed and accuracy.
Methods like TSM~\cite{tsm} and \cite{fanbuch2020rubiks} showed that a simple shift in the convolutional features results in improved accuracy at a low computational budget.

\section{Dynamic Salient Regions GNNs}
\label{sec:our_approach}

We investigate how to create node representations that are useful for modeling visual interactions between various entities in space and time using GNNs. Our proposed Dynamic Salient Regions GNN model (DyReg-GNN) learns to dynamically assign each node to a certain interesting region. By dynamic, we mean that
we have a fixed number $N$ of regions that change their position and size according to the input at each time step. The regions assigned to each of the $N$ nodes can change from one moment of time to the next depending on their saliency. 

The main architecture of our DyReg-GNN model is illustrated in Figure~\ref{fig:main_arch}. Our model receives feature volume $X~\in~\R^{T \times H\times W \times C}$ and at each time step $t$ we predict the location and size of $N$ regions. From these regions, a differentiable pooling operation creates graph nodes that are processed by a GNN and then are projected to their initial position. This module can be inserted at any intermediate level in a standard convolutional model.
%

\subsection{Node Region Generation}

We want to attend only to a few most relevant entities in the scene, thus  a small number of nodes are used in DyReg-GNN (in our experiments $N = 9$) and it is crucial to assign them to the most salient regions. The number of nodes is a hyperparameter that we choose such that it exceeds the expected number of relevant entities in the scene, to increase the robustness of the model. 
Thus, we propose a global processing (shown in Figure~\ref{fig:main_arch} B) that aggregates the entire input features to produce regions defined by parameters indicating their location $(\Delta x, \Delta y)$ and size  $(w, h)$.

To generate $N$ salient regions, we process the input $X_t$ using position-aware functions $f$ and $\{g_i\}_{i \in \overline{1,N}}$ that retain spatial information. 
Nodes should be consistent across time, thus we generate their regions in the same way at all time steps, by sharing in time the parameters of $f$ and $\{g_i\}$.
The function $f$ is a convolutional network that highlights the important regions from the input.
\begin{equation}
\label{eq:fishnet}
    M_t = f(X_t) \in \R^{H' \times W' \times C'}
\end{equation}
For each node $i$, we generate a latent representation of its associated region using the $\{g_i\}$ functions. Each $g_i$ has the same architecture, but different parameters for each node and could be instantiated as a fully connected network or as global pooling enriched with spatial positional information.
We generate the node regions from a global view to make the decision as informed as possible.
\begin{equation}
\label{eq:mlp}
    \mathbf{\hat{m}}_{i,t} = g_i(M_{t})  \in \R ^ {C'} , \forall i \in \overline{1,N}
\end{equation}
 Each of the $N$ latent representations is processed independently, with a GRU~\cite{cho_gru} recurrent network (shared between nodes),
 to take into account the past regions' representations.
\begin{equation}
\label{eq:gru}
    \mathbf{z}_{i,t} = \text{GRU}(\mathbf{z}_{i,t-1}, \mathbf{\hat{m}}_{i,t})  \in \R ^ {C'}, \forall i \in \overline{1,N}
\end{equation}
 At each time step, the final parameters are obtained by a linear projection $W_o \in \R ^ {C' \times 4}$, transformed by a function $\alpha$ to control the initialisation of the position and size (e.g. regions would start at reference points either in the center of the frame or arranged on a grid). For more details about how to set the transformation $\alpha$ we refer to the Supplemental Materials.
\begin{equation}
\label{eq:offsets}
    \mathbf{o}_{i,t} = ( \Delta x_{i,t}, \Delta y_{i,t}, w_{i,t}, h_{i,t})
    = \alpha (W_o \mathbf{z}_{i,t}) \in \R ^ 4
\end{equation}
%
\subsection{Node Features Extraction}
\label{sec:node_extraction}
The following operations are applied independently at each time step thus, in the current subsection, we ignore the time index for clarity. We extract the features corresponding to each region $i$ using a differentiable pooling w.r.t. the predicted region parameters $\mathbf{o}_i$. All the input spatial locations $p \in \R ^ 2$ are interpolated according to the  kernel function $\K^{(i)}(p)$ as presented in Figure~\ref{fig:main_arch} C.

We present the operation for a single axis since the kernel is separable, acting in the same way on both axes: 
\begin{equation}
    \K^{(i)}({p_x, p_y}) = k^{(i)}_x(p_x) k^{(i)}_y(p_y) \in \R
\end{equation}

We define the center of the estimated region $c_{i,x} + \Delta x_i$, where $c_{i,x}$ is a fixed reference point for node $i$ (located in the frame's center or arranged on a grid). The values of the kernel decrease with the distance to the center and is non-zero up to a maximal distance of $w_i$, where $w_i$ and $\Delta x_i$ are the predicted parameters from Eq.~\ref{eq:offsets}.
\begin{equation}
     k^{(i)}_x(p_x) = \text{max}(0, w_i - | c_{i,x} + \Delta x_i - p_x | )
\end{equation}
For each time step $t$, node $i$ is created by interpolating all points in the input $X_t$ using the kernel function. By modifying $(\Delta x_i,  \Delta y_i)$ the network controls the location of the regions, while $(h_i, w_i)$ parameters indicate their size.
\begin{equation}
    \mathbf{v}_{i} = \sum_{p_x=1}^{W}\sum_{p_y=1}^{H}  \K^{(i)}(p_x, p_y) \mathbf{x}_{ p_x,p_y}  \in \mathbb{R}^{C}
\end{equation}
Setting $w_i=1$ leads to standard bilinear interpolation, but optimising it allows the model to adapt region's size and we observe that larger ones result in a more stable optimisation (see node size ablations from Supp. Material).

The position of the region associated with each node should be taken into account. It helps the relational processing by providing an identity for the node and is also useful in tasks that require positional information. We achieve this by computing a positional embedding for each node $i$ using a linear projection of the kernel $\K_i$ into the same space as the feature vector $v_i$ and summing them.



\paragraph{Key Properties.}

By construction, the nodes in our method are \textit{localized}, meaning that they are clearly associated with a location: they pool information from clearly delimited area in space and they maintain position information from the positional embedding. These two aspects could be helpful in tasks involving spatio-temporal reasoning.

The \textit{dynamic} aspect refers to the key capability of adapting the region's position and size according to the saliency of the input at each time step. This is done by predicting the regions from the input with the operations from \cref{eq:fishnet,eq:mlp,eq:gru,eq:offsets}.



An essential aspect of this method is that the final classification loss is \textit{differentiable} with respect to regions' parameters as the gradients are passing from the nodes outputs $v_{i}$ through the kernels $k_i$ to the parameters $w_i$ and $\Delta x_i$. 
This allows us to learn regions from the final loss, \textit{without direct supervision for the region generation}. Thus the method has more flexibility in learning relevant regions as appropriate for the task.


\subsection{Graph Processing}
\label{graph-processing}

For processing the nodes' features, different spatio-temporal GNNs could be used. Generally, they follow a framework~\cite{pmlr-v70-gilmer17a} of  sending messages between connected nodes, aggregating~\cite{velickovic2018graph_gat, xu2018how_powerful} and updating them.

The specific message-passing mechanism is not the focus of the current work, thus we follow a general formulation similar to \cite{rstg} for recurrent spatio-temporal graph processing. It uses two different stages: one happening between all the nodes at a single time step and the other one updating each node across time. For each time step $t$, we send messages between each pair of two nodes, computed as an MLP (with shared parameters) and aggregates them using a dot product attention $a(v_i,v_j) \in \R$.
\begin{align}
    \mathbf{v}_{i,t} &=   \sum_{j=1}^{N} a(\mathbf{v}_{j,t},\mathbf{v}_{i,t}) \text{MLP} ([\mathbf{v}_{j,t};\mathbf{v}_{i,t}])  \in \mathbb{R}^{C}
\end{align}
We incorporate temporal information through a shared recurrent function across time, applied independently for each node.

\begin{equation}
    \mathbf{\hat{v}}_{i,t+1} = \text{GRU}(\mathbf{\hat{v}}_{i,t}, \mathbf{v}_{i,t}) \in \mathbb{R}^{C}
\end{equation}
The GRU output represents the updated nodes' features and the two steps are repeated $K=3$ times. 

\subsection{Graph Re-Mapping} 
%
To use our method as a module inside any backbone, we produce an output with the same shape as the convolutional input $X_t \in~\R^{H\times W \times C} $. The resulting features of each node are sent to all locations in the input according to the weights used in the initial pooling from Section~\ref{sec:node_extraction}.

%
\begin{equation}
    \mathbf{y}_{p_x,p_y,t} = \sum_{i=1}^{N}  \K^{(i)}_t(p_x, p_y) \mathbf{\hat{v}}_{i,t}  \in \mathbb{R}^{C}
\end{equation}
\section{Experimental Analysis}
While much effort is put into the creation of different video datasets used in the literature, such as Kinetics~\cite{carreira2017quo} or Charades~\cite{sigurdsson2016hollywood_charades}, it has been argued~\cite{cater} that they contain biases that make them solvable without complex spatio-temporal reasoning. CATER~\cite{cater} is proposed to alleviate this, but it is too small (5500 videos) and still has biases that make the last few frames sufficient for good performance~\cite{hopper}.
We test our model on two video classification datasets that seem to offer the best advantages, being large enough and requiring abilities to model complex interactions.
We evaluate on real-world datasets, Something-Something-V1\&V2~\cite{goyal2017something}, while we also test on a variant of the SyncMNIST~\cite{rstg} dataset that is challenging and requires spatio-temporal reasoning, while allowing fast experimentation.  The code for our method can be found in our repository \footnote{\url{https://github.com/bit-ml/DyReg-GNN}}.

\subsection{Human-Object Interactions Experiments}
Something-Something-V1\&V2~\cite{goyal2017something} datasets classify scenes involving human-object complex interactions. They consist of 86K / 169K training videos and 11K / 25K validation videos, having 174 classes. Unless otherwise specified, all experiments on Something-Something datasets use TSM-ResNet-50~\cite{tsm} as a backbone and we add instances of our module at multiple stages.

\begin{figure}[t]
\begin{minipage}{0.59\textwidth}
\small
\captionof{table}{
Results on val. set of Smt-Smt-V2 showing the \textbf{importance of salient regions discovery.}
We compare our predicted (unsupervised) regions to fixed grid regions or boxes given by an object detector using the same GNN model.
The mean $L_2$ distance between the regions and gt. objects proves that DyReG-GNN has regions correlated with objects, while also having superior accuracy and efficiency.
}
\begin{tabular}{l c r c c}
     \cmidrule[1pt](lr){1-5}
    Model & Regions & FLOPS & Dist & Acc\\
    & discovery & \multicolumn{1}{c}{$\downarrow$} & $\downarrow$ & (\%)$\uparrow$ \\

    \cmidrule(lr){1-5}
    TSM-R50 & - & 65.8G & - & 63.4 \\
    \cmidrule(lr){1-5}
     ~+ GNN+Fixed & Grid & +1.4G  & 0.170 & 64.1\\
     ~+ GNN+Detector & Obj detector & +41.1G & 0.125  & 64.0 \\
     ~+ DyReg-GNN & Unsupervised & +1.6G & 0.129 & \textbf{64.8}    \\
\end{tabular}
\label{tab:object_oriented2}
\vspace{-4mm}
\end{minipage}
\hfill
\begin{minipage}{0.39\textwidth}
\small
\captionof{table}{ 
\textbf{Consistent improvements over different backbones} on the validation set of Smt-Smt-V1 using central crop evaluation.
}
\begin{tabular}{l l c c c}
     \midrule
    Model &  \multicolumn{1}{c}{Acc (\%)} \\
    \midrule
    TSM-R18 & 33.7 \\
    TSM-R18 + DyReg-GNN & 35.6 ($\uparrow 1.9$) \\
    \midrule
    I3D-R50 & 44.0 \\
    I3D-R50 + DyReg-GNN & 45.4 ($\uparrow 1.4$)\\
    \midrule
    TSM-R50 & 	47.2\\
    TSM-R50 + DyReg-GNN & 48.8 ($\uparrow 1.6$)\\

\end{tabular}
\label{tab:backbone}
\vspace{-4mm}
\end{minipage}
\end{figure}

\smallskip\noindent\textbf{Studying the Importance of Salient Regions Discovery.}
\phantomsection \label{para:smt_salient_reg}
We test the importance of the dynamic regions for GNNs vision methods by training models where we replace the predicted regions with the same number of fixed regions on a grid (GNN + Fixed Regions) or boxes (GNN + Detector) as given by a Faster R-CNN~\cite{ren2015faster_rcnn} trained on MSCOCO~\cite{lin2014coco}. 


%

The detector based model has comparable results to the one with fixed regions, seemingly being unable to fully benefit from the correctly identified objects. 
The relative weaker performance of this model could be due to the fact that the pre-trained detector is not well aligned to the actual salient regions that are relevant for the 
classification problem. 

On the other hand, this weakness is not applicable for DyReg-GNN that learns suitable regions for the current task and it obtains the best performance as seen in Table~\ref{tab:object_oriented2}.
Not only that it does not require object annotations, but it is also more computationally efficient. Running the detector on a video of size $224 \times 224$ would add $39.7$ GFLOPS on its own, comparing to the $1.6$G of three DyReg-GNN modules, from which $0.2$G represents the regions prediction.

Overall, our method, with unsupervised regions obtains superior performance in terms of accuracy and computational efficiency representing a suitable choice for relational processing of a video.




\smallskip\noindent\textbf{Object-centric representations.}
\phantomsection \label{para:smt_object_centric}
The nodes represent the core processing units and their localization enforces a clear decision on what specific regions to focus on while completely ignoring the rest, as a form of hard attention. Different from other works~\cite{show_attend_tell}, our hard attention formulation is differentiable. To better understand what elements influence the model predictions, we could inspect the predicted kernels, thus introducing another layer of interpretability to the model, on top 
of the capabilities offered by the convolutional backbone.
Visualisations of our nodes' regions reveal that generally, they cover the objects in the scene. 
For example, in the first row of Figure~\ref{fig:qualitative} the nodes are placed around the phone in the first frames and then separate into two groups, one for the phone one for the hand. \looseness=-1


The localized nodes make our model capable of discovering salient regions, leading to object-centric node representations.
We quantify this capacity by measuring the mean $L_2$ distance (normalised to the size of the input) between the predicted regions and ground-truth (gt.) objects given by \cite{CVPR2020_SomethingElse}. The metric is completely defined in the Supp. Materials.
We observed that the score improves during the learning process (it reaches $0.129$  starting from $0.201$), although the model is not optimized for this task. This suggests that the model actually learns object-centric representations.

In Table~\ref{tab:object_oriented2} we also compare the final $L_2$ distance of our best DyReg-GNN model to an object detector and to fixed grid regions. 
Although our method is not designed and supervised to find object regions, we observe that it is able to predict locations that are fairly close to gt. objects. The $L_2$ distance is similar to the one obtained by an external model ($0.129$ vs $0.125$), trained especially for detecting objects. \looseness=-1


We observe that learning the regions' size is important for the stability of the optimisation and thus for the final performance (see Tab.\ref{tab:mnist_ablations} and Supp. Material -  Regions’ Size section). However, the predicted size is not as well aligned with the size of the true objects. This gives us a hint that for the action classification task it is important to have good region locations, but their size is less relevant.
We leave a more thoroughly investigation for futures work.

These experiments prove that the high-level classification task is well inter-related with the discovery of salient regions and that, in turn, these regions improve the relational processing in the recognition task.
First, we show that DyReg-GNN's region obtain superior accuracy and efficiency than other methods of extracting nodes and second, these regions are well correlated to gt. object locations.

\begin{table*}[t!]
\parbox{.51\linewidth}{
\small
\centering
\caption{ \textbf{Results on val. set of Smt-Smt-V1.} Our model achieves competitive results compared to recent works (best results in \textcolor{blind1_red}{red}), while it outperforms all other graph-based methods (best results in \textcolor{blind1_blue}{blue}). All the methods use ResNet50 as backbone.
}

\begin{tabular}{p{0.5mm} l l c c c l l}
    \cmidrule[1pt](lr){1-6}
    & Model & Regions & \#F & Top 1 & Top 5 \\
    & & discovery & & & \\
     \cmidrule(lr){1-6}
    \multirow{6}{*}{\rotatebox[origin=c]{90}{non-Graph}} 
    & TSM~\cite{tsm} & - & 16 & 48.4 & 78.1 \\
    & S3D~\cite{xie2018rethinking} & - & 64  & 48.2 & 78.7 \\
    &GST~\cite{luo2019GST} & - & 16  & 48.6 & 77.9 & \\
    &SmallBig~\cite{Li2020SmallBigNetIC} & - & 16 & 50.0 & 79.8  \\
    &STM~\cite{jiang2019stm} & - & 16 &  50.7 & 80.4  \\
    &MSNet~\cite{motionsqueeze_eccv2020} & - & 16 & \textbf{\color{blind1_red} 52.1} & \textbf{\color{blind1_red} 82.3} \\
    \cmidrule(lr){1-6}
    \multirow{6}{*}{\rotatebox[origin=c]{90}{Graph}} & ORN~\cite{Baradel_2018_ECCV_orn} & Objects & 8  & 36.0 & - \\
    & NL I3D~\cite{wang2018videos_gupta2} & Grid & 32 & 44.4 & 76.0 \\
    & NL GCN~\cite{wang2018videos_gupta2} & Objects & 32  & 46.1 & 76.8 \\
    & TRG~\cite{Zhang2020TemporalRG} & Frames & 16  &  48.1 & \textbf{\color{blind1_blue} 80.4}  \\
    & RSTG~\cite{rstg} & Grid & 32 & 49.2 & 78.8 \\
    \cmidrule(lr){2-6}
    & TSM+DyReg & Dynamic & 16  &  \textbf{\color{blind1_blue} 49.9} & 79.0 \\

\end{tabular}
\label{tab:smt-smt-v1-results}
\vspace{-5mm}
}
\hfill
\parbox{.45\linewidth}{
\small
\centering
\caption{\textbf{Results on val. set of Smt-Smt-V2.}, in comparisons to recent works. DyReg-GNN improves the TSM-ResNet50 backbone when using either one (r4) or three (r3-4-5) modules of graph processing and it obtains top results.
}
\begin{tabular}{l l l l l l}
    \cmidrule[1pt](lr){1-5}
    Model & BB & Top 1 & Top 5 \\
    
    \cmidrule(lr){1-5}
    
    %
    %
    TRG~\cite{Zhang2020TemporalRG} & R50 &  59.8 & 87.4 \\
    
    GST~\cite{luo2019GST} & R50 &  62.6 &87.9  & \\
    v-DP~\cite{zhou2020spatiotemporal} & D121 &  62.9 & 88.0 \\
    SmallBig~\cite{Li2020SmallBigNetIC} & R50 &  63.8 & 88.9  \\
    STM~\cite{jiang2019stm} & R50 &  64.2 & \textbf{89.8} \\
    MSNet~\cite{motionsqueeze_eccv2020} & R50 &  64.7 & 89.4 \\
    TSM~\cite{tsm} & R50 &  63.4 & 88.5 \\
    \cmidrule(lr){1-5}
    TSM+DyReg-r4 & R50 &  64.3 & 88.9 \\
    TSM+DyReg-r3-4-5 & R50 &  \textbf{64.8} & 89.4 \\
\end{tabular}
\label{tab:smt-smt-v2-results}
\vspace{-6mm}
}
\end{table*}
\smallskip\noindent\textbf{Comparison to recent methods.} 
\phantomsection \label{para:smt_sota}
DyReg-GNN can be used with any convolutional model and we show that it consistently boosts the performance of multiple backbones(Table~\ref{tab:backbone}). We compare to recent methods from the literature in Table~\ref{tab:smt-smt-v1-results} and Table~\ref{tab:smt-smt-v2-results}. 
Our method improves the accuracy over the TSM-ResNet50 backbone on both Smt-Smt-V1 and Smt-Smt-V2 by $1.5\%$  and $1.4\%$ respectively and achieves competitive results.
Compared to all the other graph based methods we obtain superior results, showing that our discovery of dynamic regions is effective for space-time relational processing. \looseness=-1






\smallskip\noindent\textbf{Implementation Details}
Unless otherwise specified, we use TSM-ResNet50 (pre-trained on ImageNet~\cite{imagenet}) as our backbone and add instances of our module in the last three stages. To benefit from ImageNet pre-training, we add our graph module as a residual connection. We noticed that models using multiple graphs have problems learning to adapt the regions from certain layers. We fix this by training models containing a single graph at each single considered stage, as the optimisation process is smoother for a single module, and distill their learned offsets into the bigger model. 
The distillation is done for the first $10\%$ of the training iterations to kick-start the optimization process and then continue the learning process using only the video classification signal.

In all experiments we follow the training setting of \cite{tsm}, using $16$ frames resized to have the shorter side of size $256$, and randomly sample a crop of size $224 \times 224$. 
%
For the evaluations, we follow the setting in \cite{tsm} of taking 3 spatial crops of size $256 \times 256$ with 2 temporal samplings and averaging their results. For training, we use SGD optimizer with  learning rate $0.001$ and momentum $0.9$, using a total batch-size of $10$, trained on two GPUs. We decrease the learning rate by a factor of 10 three times when the optimisation reaches a plateau.

\vspace{-3mm}
\subsection{Synthetic Experiments}
\label{sec:synthetic_experiments}
\vspace{-2mm}
SyncMNIST is a synthetic dataset involving digits that move on a black background, some in a random manner, while some move synchronously. The task is to identify the digits that move in the same way.
We use a harder variant of the dataset (MultiSyncMNIST), where the videos could include multiple digits of the same class.
The challenge consists in finding useful entities and model their relationships while being able to distinguish between instances of the same class. 
Each video contain 5 digits and the goal is to find the smallest and the largest digit class among the subset that moves in the same way. This results in a video classification task with 56 classes. The dataset contains 600k training videos and 10k validation videos with 10 frames each.


%
\vspace{-3mm}
\paragraph{Studying the Importance of Dynamic Nodes.} 
\phantomsection \label{para:mnist_dynamic_importance}
%
We validate our assumption that the nodes should be dynamic, meaning that their regions position and size should be adapted according to the input at each time step.
We investigate (Table.~\ref{tab:mnist_ablations}) different types of localized nodes, each adapting to the input to a varying degree, and show the benefits of our design choices. We experiment with variants of our model, all having the same backbone (2D ResNet-18~\cite{resnet}), the same graph processing and same pre-determined number of regions, but we constrain the node regions in different ways.

\textit{Fixed Model} extracts node features from regions arranged on a grid, with a fix location and size.

\textit{Static Model} investigates the importance of dynamic regions by optimising regions based on the whole dataset but do not take into account the current input.
Effectively, the features $\mathbf{z}_i$ from Eq.~\ref{eq:offsets} become learnable parameters.

\textit{Constant-Time Model} has regions adapted to the current video but they do not change in time.

\textit{DyReg-GNN Model} predicts regions defined by location and size, and we can either pre-determine a fixed size for all the regions (\textit{Position-Only Model}) or directly predict it from the input as in our complete model (\textit{DyReg-GNN Model}). 

\begin{figure}[t]
\begin{minipage}{0.55\textwidth}
\centering
     \includegraphics[width=0.98\textwidth]{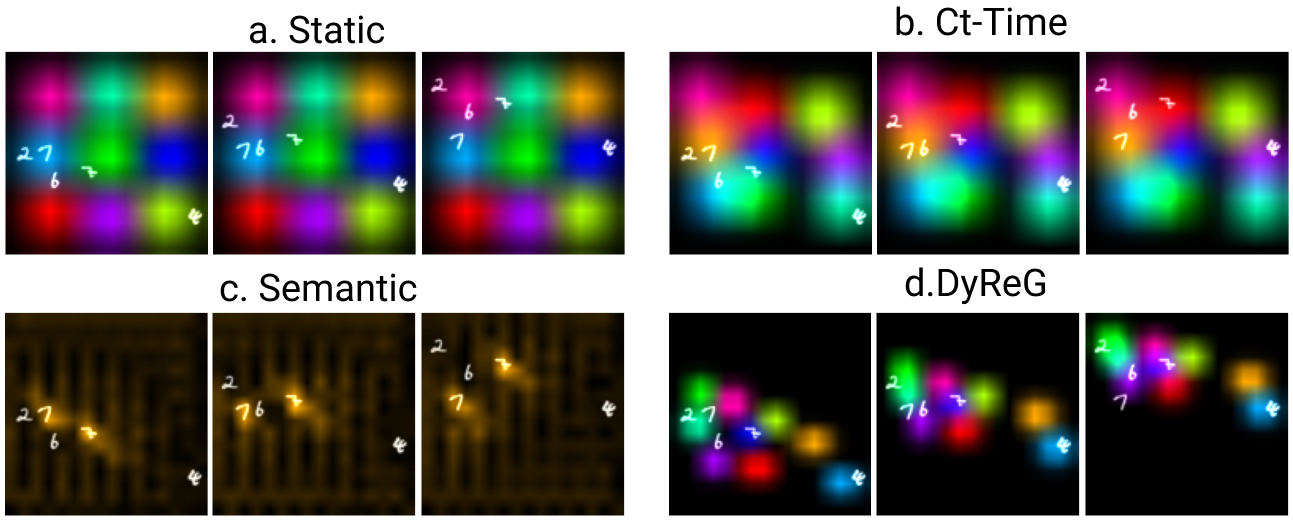}
     \caption{
     \textbf{Nodes' regions} on MultiSyncMNIST for 3 frames.
     a) \textit{Static} Model, ignoring the input, learns a regular grid; 
     b) \textit{Constant-Time} predicts the same regions for all time steps, covering the movement in the video;
     c) The attention map of a single node of \textit{Semantic} that can't distinguish between different instances of the same digit;
     d) \textit{DyReg-GNN} generally follows the digits locations at each time steps while also adapting the regions' size.
     \vspace{-4mm}
     }
     \label{fig:ablations_viz}
\end{minipage}
\hfill
\begin{minipage}{0.42\textwidth}
\centering
\includegraphics[width=1.0\textwidth]{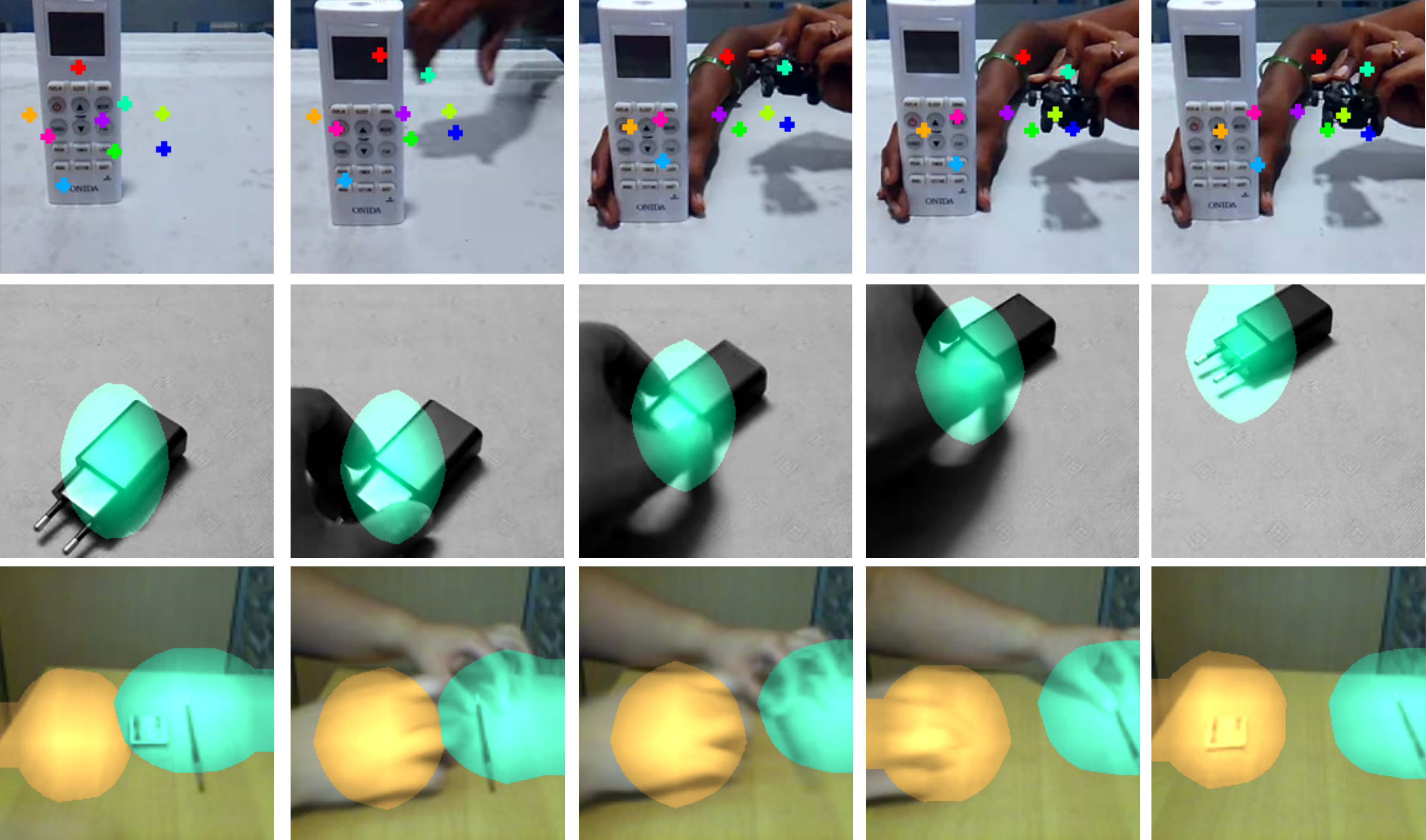}
\captionof{figure}{
\textbf{Nodes' regions} on Smt-Smt-V2.
We show (1st row) the center of all the $N$ regions as predicted by DyReg-GNN (each color for a node). Each node region (last 2 rows) corresponds to a zone from the \textit{latent} conv features pooled by a node.
\vspace{-4mm}
}
\label{fig:qualitative}
\end{minipage}
\end{figure}

These experiments (Table~\ref{tab:mnist_ablations}), show that the fixed region approach (Fixed Model) achieves the worst results, slightly improving when the regions are allowed to change according to the learned statistics of the dataset (Static model).
Adapting to the input is shown to be beneficial,
the performance improving even when the regions are invariant in time (Constant-Time Model), and further more when predicting different regions at every time steps (Position-Only). The best performance is achieved when both the location and the size of the regions are dynamically predicted from the input (DyReg-GNN). \looseness=-1

In Figure~\ref{fig:ablations_viz} we show examples of the kernels obtained for each of these models. We observe that the Static Model's kernels are learned to be arranged uniformly on a grid, to cover all possible movements in the scene, while the Constant-Time Model's kernels are adapted for each video such that they cover the main area where the digits move in the current video. The full DyReg-GNN Model learns to reduce the size of its regions and we observe that they closely follow the movement of the digits.



The previous experiments show that performance increases when the model becomes more dynamic, proving that our model benefits from nodes that are adapted to a higher degree to the current input.\looseness=-1
\vspace{-2mm}
\paragraph{Studying the Importance of Localized Nodes.}
\phantomsection \label{para:mnist_localized_importance}
We argue that nodes should pool information from different locations according to the input, such that the extracted features correspond to meaningful entities. Depending on the goal, we could balance between semantic nodes globally extracted from all spatial positions or localized (spatial) nodes that are obtained from well-delimited regions.

\textit{Semantic Model} creates nodes similar to \cite{glore, sgr_symbolic_graph} where each node extracts features from all the spatial locations and could represent a semantic concept.
Each node is extracted by a global average pooling where the weights at every position $p$ are directly predicted from the input features at that location. Practically, we replace the spatially delimited kernel used in our model with this global attention map.

A major downside of this approach is that it does not distinguish between positions with the same features, making it harder to reason about different instances. 
Figure~\ref{fig:ablations_viz}.C shows the attention map of a single node and we observe that it has equally high activations for both instances of the same digit, thus making it hard to distinguish between them.



%

This limitation does not exist in our DyReg-GNN model, as it predicts localized nodes that favour the modeling of instances. For comparison, we use two variants with a different number of parameters and show that they clearly outperform the semantic model (Table~\ref{tab:mnist_results}).
%
These experiments prove that in cases that involve spatial reasoning of entities, such as the current task, DyReg-GNN is a perfect choice, showing its benefits for spatio-temporal modeling.

\begin{table*}
\parbox{.57\linewidth}{
\caption{\textbf{Ablation of dynamic nodes on MultiSyncMNIST.} It is crucial to have regions that adapt based on the input (Dynamic), both their position (Pos.) and size at each time step.
}
\centering
\small
\begin{tabular}{l  c  c  c  c  c}
    \toprule
    Model & Optimise & Time & \multicolumn{2}{c@{}}{Dynamic}  & Acc 	\\
     & Pos. & Varying & Pos. & Size &  \\
    \midrule
    Fixed & & & & & 78.85 \\
    Static & \checkmark & & & & 81.48 \\
    Ct-Time & \checkmark &  & \checkmark & & 86.77 \\
    Pos-Only & \checkmark & \checkmark & \checkmark & & 93.41 \\
    DyReg-GNN & \checkmark & \checkmark & \checkmark& \checkmark & \textbf{95.09}\\
\end{tabular}
\label{tab:mnist_ablations}
\vspace{-4mm}
}
\hfill
\parbox{.40\linewidth}{
\centering
\caption{ \textbf{Semantic vs spatial nodes} on MultiSyncMNIST. The localized (spatial) node regions of DyReg-GNN are better suited than semantic nodes' maps obtained by the Semantic Model.
}
\small
\begin{tabular}{l c c}
    \toprule
    Model & Params (M) & Acc \\
    \midrule
    ResNet-18 & 2.79 & 52.29\\
    Fixed & 2.82 & 78.85 \\
    Semantic & 2.85 & 82.41\\
    \midrule
    DyReg-GNN-Lite & 2.83 &  91.43  \\
    DyReg-GNN & 3.08 &  \textbf{95.09}  \\
\end{tabular}
\label{tab:mnist_results}
\vspace{-4mm}
}
\end{table*}


\looseness=-1\smallskip\noindent\textbf{Implementation details.}
All models share the ResNet-18 backbone with 3 stages, 
where the graph receives the features from the second stage and sends its output to the third stage. We use $N = 9$ graph nodes and repeat the graph propagation for three iterations. In our main model, $f$ from Eq.~\ref{eq:fishnet} is a small convolutional network while $g$ is a fully connected layer. For the lighter model  that implements $g$ as a global pooling enriched with spatial positional information, we refer to the Supp. Materials. The graph offsets are initialized such that all the nodes' regions start in the center of the frame. In all experiments, we use SGD optimizer with learning rate $0.001$ and momentum $0.9$, trained on a single GPU.

\smallskip\noindent\textbf{Key Results.}  In the previous section, we experimentally validated that: \textbf{1.} DyReg-GNN consistently improves multiple backbones (Table~\ref{tab:backbone}) obtaining competitive results (Table~\ref{tab:smt-smt-v1-results},~\ref{tab:smt-smt-v2-results});
\textbf{2.} learned dynamic regions are crucial for good performance (Table~\ref{tab:mnist_ablations}) and \textbf{3.} these regions are preferable to fixed regions or external object detectors for space-time GNNs (Table~\ref{tab:object_oriented2}); 
%
%
\textbf{4.} predicted nodes correspond to salient regions (Fig.~\ref{fig:ablations_viz}-\ref{fig:qualitative}) and are well correlated with objects (Table~\ref{tab:object_oriented2}). 

\vspace{-2mm}
\section{Conclusions}

We propose Dynamic Salient Regions Graph Neural Networks (DyReg-GNN), a relational model 
for processing spatio-temporal data (videos), that augments visual GNNs by learning to predict localized nodes, adapted for the current scene. 
This novel method enhances the relational processing of spatio-temporal GNNs and we experimentally prove that it is superior to having nodes anchored in fixed predefined regions or linked to external pre-trained object detectors. 
Although we do not use region level supervision, the learning dynamics of high-level classification produces salient regions that are well correlated with object instances. 
%
%
We believe that our method of learning dynamic, localized nodes is a valuable direction that could lead to further advances to the growing number of powerful relational models in spatio-temporal domains.

\paragraph{Acknowledgment}

We would like to thank Florin Brad, Elena Burceanu and Florin Gogianu for their valuable feedback and discussions of this work. This work has been supported in part by Bitdefender and UEFISCDI, through projects EEA-RO-2018-0496 and PN-III-P4-ID-PCE-2020-2819.





\bibliography{arxiv}
\bibliographystyle{unsrt}

\appendix
\newpage
\section*{\centering\LARGE\bf Appendix: Discovering Dynamic Salient Regions for Spatio-Temporal Graph Neural Networks}
\vspace{5mm}


In this Appendix we present an impact statement, discuss some limitations of the method and then we provide more technical details about DyReg-GNN model and include some additional visualisations and ablation studies. 

Section~\ref{sec:impact} presents some views on the broader impact of this work.

Section~\ref{sec:limitations} identifies some limitations of the methods.

Section~\ref{sec:node_gen} presents more details about how the regions are generated.

Section~\ref{sec:visualisation} shows a qualitative analysis of the regions predicted by our model.

Section \ref{sec:multi_sync} shows  additional ablation studies in the synthetic setting in relation to the number of nodes, the regions size, the importance of recurrence when generating the nodes and comparisons to using ground-truth boxes or other baselines.

Section~\ref{sec:smt-smt} presents 
some training details, describe the metric used to measure the correlation between our regions and the existing objects in the scene and have a runtime analysis of our proposed module.


We provide our full code as supplementary material and we will release it online upon the paper publication. Beside this Appendix, we also provide some videos, visualising the regions discovered by our DyReg-GNN model.


\section{Broader Impact}
\label{sec:impact}

 
We research novel methods that would improve current general models for spatio-temporal processing. Our goal is to investigate models that  emphasize a small number of relevant nodes having the potential to be more explainable and that could lead to more interpretable reasoning. Although this is not fully realised in this paper, we believe that this work is a good step in this direction. 
Our model enhances any convolutional backbone for video processing and thus inherits the benefits and also the possible harms brought by such models. 

When developing our model, we used a synthetic dataset of moving digits and a public dataset for human-object interactions. Our model is kept generic, with no parts specially designed for these tasks. The models trained on these datasets have no obvious direct real application, as the first one is a toy dataset and the second one has restrictive classes meant only to evaluate the capabilities of the models. But developing better models for video understanding leads to more effective applications. On one hand, it could lead to better applications helping visually impaired people navigate the world and on the other hand it could lead to stricter automatic surveillance of workers. 
In order for ML technology to have a positive broader impact, more discussions between different actors in society should be conducted leading to the development of guidelines and practices.

The proposed work does not rely on using object detectors and only uses video level supervision. Object detectors have a predefined list of objects, that would not be sufficient for many practical cases leading to biases in the system. Moreover, this way we eliminate a possible source of biases coming from the object-level annotations.




\section{Limitations}
\label{sec:limitations}

By design, DyReg-GNN uses a fixed number of nodes, that we treat as a hyperparameter. This way the model is forced to produce the same number of regions regardless of the complexity of the scene. From simpler scenes, the model learns to group the nodes in overlapping regions, creating redundancy. On the other hand, more complex scenes have an increased number of relevant regions, tending to require distinct regions. This could lead to a discrepancy that would increase the difficulty of the optimisation process.
Changes in scene's complexity could be also observed in a single video when the scene suffer major changes in time. For example when elements appear, disappear or are occluded from view, the number of regions predicted by the model remains the same and it is harder to properly model all the elements.
Ideally, we want a system that adapts to the complexity of the scene by dynamically predicting the number of nodes. This is a challenging task that requires additional investigations and we leave it for future work.

Preliminary experiments reveals that our method requires a relative large amounts of data to be properly trained. This seems not to be an issue for Something-Something dataset that has 80k-160k training videos, but could be an issue for smaller datasets. On MultiSyncMNIST we could train models with high accuracy on $10\%$ of the whole dataset of 600k videos. But when using only $1\%$ (6k videos) of the data, the predicted regions would not change during training. Given the size of the recent video dataset, this is not a big limitation.



\section{Node Region Generation}
\label{sec:node_gen}


The goal of this sub-module is to generate the regions that correspond to salient zones in the input. We achieve this by processing the input globally with position-aware functions $f$ and $\{g_i\}$. 

\paragraph{Function $f$.} We use $f$ function to aggregate local information from larger regions in the input while preserving sufficient positional information. The input $X_t \in \R^{H~\times W~\times C}$ is first projected into a lower dimension $C'$ since this representation should only encode saliency without the need to precisely model visual elements. Then we increase the receptive field by applying two conv layers, followed by a transposed conv and then a final conv layer. This results in a feature map  $M_t = f(X_t) \in \R^{H' \times W' \times C'}$.
Depending on the backbone and the stage where the graph is added $H, W$ have different values and we adapt the hyperparameters of the convolutional layers such that $H'$ and $W'$ are not smaller than $6$. For example, in the synthetic experiments $f$ reduces the input from $\R^{16 \times 16 \times 32}$ to $\R^{7 \times 7 \times 16}$.

\paragraph{Functions $\{g_i\}$.} For each node $i$ we use $g_i$ to extract a global latent representation from which we predict the corresponding region parameters. We present two variant of $g_i$ function, a larger and more precise one and a smaller, more computational efficient one. 

For the bigger one, we use a simple fully connected layer of size $C \times (H' * W' * C') $ that takes the whole $M_t$ and produces a vector of size $C$. This way $g_i$ could distinguish and model the spatial locations of the $H' \times W'$ grid.

The second approach consists in a weighted global average pooling for each node $i$. The weight associated to each location $p$ is predicted directly from the input $M_{t,p}$ by a $1 \times 1$ convolution. But this results in a translation-invariant function $g_i$ that losses the location information. We alleviate this by adding to each of the $H' \times W'$ location a positional embedding similar to the one used in \cite{vaswani2017attention}. This approach predicts regions of slightly poorer quality as the location information is not perfectly encoded in the positional embeddings. For a lighter model, such as the one presented in Table~\ref{tab:mnist_results} of the main paper we could use the second approach for the $\{g_i\}$ functions and also skip the $f$ processing.

\paragraph{Constraints}
Equation~\ref{eq:offsets} in the main paper could be expended as:

\begin{align}
    \tilde{\mathbf{o}}_{i} &= ( \Delta \tilde{x}_{i}, \Delta \tilde{y}_{i}, \tilde{w}_{i}, \tilde{h}_{i}) =  \gamma \odot W_o \mathbf{z}_{i} \in \R ^ 4 \\
    \mathbf{o}_{i} &= \alpha(\tilde{\mathbf{o}}_{i})\notag
\end{align}

To constrain the model to predict valid image regions and also to start from regions with favourable position and size, we apply non-linear functions for each component
$\mathbf{o}_{i} = \alpha(\tilde{\mathbf{o}}_{i})$
. We design the non-linearities such that $w_i, h_i > 0$ and $ \Delta x_{i} + C_x \in [0,W]$ and , $\Delta y_{i} + C_y \in [0,H]$, where $C$ is a fixed reference point. In experiments, all nodes share the same constant $C$ , representing the center of the image.
\begin{align}
    h &= e^{\tilde{h}} h_{init}  \qquad \qquad  w = e^{\tilde{w}} w_{init} \qquad \quad \\
    \Delta  {x} = \frac{W}{2} &\tanh{\Big( \Delta \tilde{x} + 
    \arctanh{( \frac{2C_x}{W} - 1)} \Big)} + \frac{W}{2}  - C_x \\
    \Delta  {y}  = \frac{H}{2} &\tanh \Big( \Delta \tilde{y} + \arctanh( {\frac{2C_y}{H} - 1)}  \Big) + \frac{H}{2} - C_y
\end{align}

By initialising $\gamma = 0$ we obtain $h = h_{init}$, $w = w_{init}$ and $\Delta  y = \Delta x = 0$. This means that all regions are initialized centered in the reference point $C$ and start with the predefined size. By default we set $h_{init}=\frac{H}{6}$, $w_{init}=\frac{W}{6}$.

\begin{figure}[t]
\begin{minipage}{0.5\textwidth}
\centering
\includegraphics[width=0.95\textwidth]{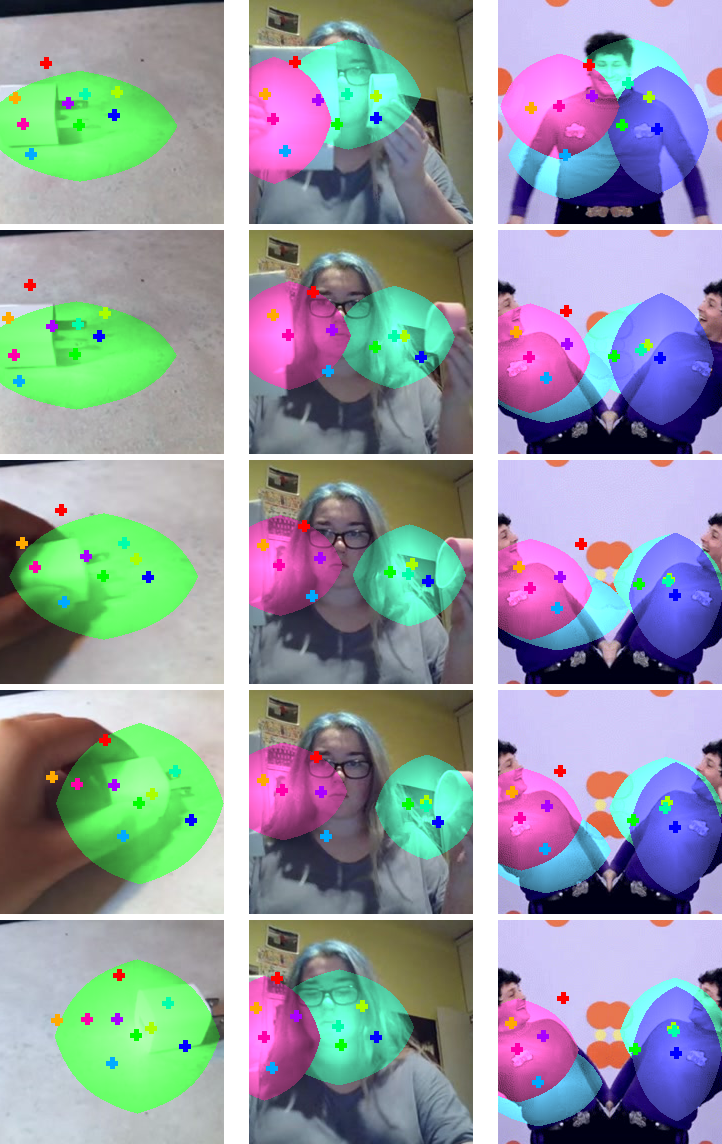}
\captionof{figure}{Visualisations of salient regions associated with each node, as predicted by our DyReg-GNN model on videos from Smt-Smt-v2 dataset (Left and Center) and on out-of-distribution real-world videos (Right).
Each node learns to move to different relevant regions in the input. For each video, we show the centers corresponding to all the nodes and, for a better visualisation, a subset of the predicted regions. Due to the receptive field of the backbone, the nodes are actually influenced by larger regions in the initial input.}
\label{fig:qualitative_regions}
\end{minipage}
\hfill
\begin{minipage}{0.47\textwidth}
\centering
\includegraphics[width=1.0\textwidth]{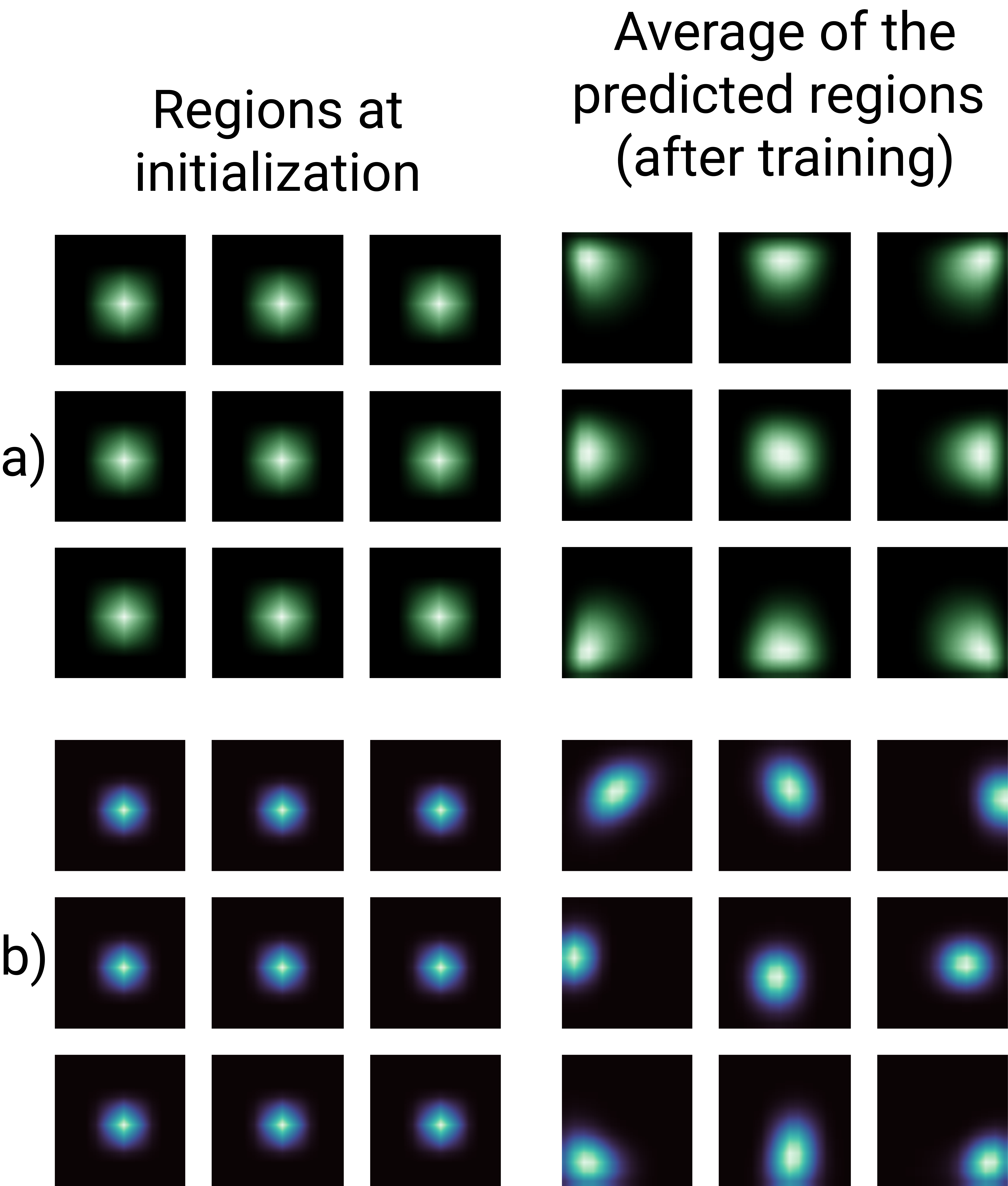}
\captionof{figure}{Visualisation of the average locations of the salient regions associated with DyReg-GNN's nodes, computed over the validation set of (a) MultiSyncMNIST  and (b) Smt-Smt-v2.
In these visualisations the order of the regions is manually selected. In the left column we show the regions at initialisation, and in the right column we present the mean regions as predicted by our learned DyReg-GNN model.
Here, we keep the size of the regions fixed, each node has a preferred location in space and assigns salient regions around it. This behaviour is learned by the model to break the symmetry of the nodes.
}
\label{fig:mean_regions}
\end{minipage}
\end{figure}

\section{Visualising the nodes' regions}
\label{sec:visualisation}

The region associated with each node is clearly delimited in space and we can easily visualize them. 
We train a model 
on Something-Something-V2 dataset of human-object interactions and in Figure.\ref{fig:qualitative_regions} we show its predicted nodes' regions for two videos from the dataset and one out-of-distribution video. Generally the nodes follow relevant regions in the input. 
We note that the visualisation of the regions is only an approximation of the actual regions that send information to the graph nodes. 
Each node pools info from a low-resolution region in the latent convolutional features, that corresponds to the high-resolution visualized region. But, the actual area that contributes to each node is actually larger, due to the receptive field of the convolutional network. Moreover, the backbone also contains temporal processing (e.g. in the form of temporal feature shifting in the case of TSM or 3D conv for I3D) such that each node receives information from adjacent time steps. Thus, we expect some misalignment in the visualizations both in space and time.

To better understand how each node attends to the input, we compute the average of its associated regions over the entire evaluation dataset (see Figure.~\ref{fig:mean_regions}).
We observe that the regions are initialized in the center of the image and, after training, each node learns to attend to regions around a specific location. For each video, a node predicts a different region, according to the input, but it is situated mostly around a certain part of the image. 
This behaviour is learned by the model to break the symmetry of the nodes and be able to create an implicit matching between relevant parts of the input and the nodes. 

\section{Synthetic Setting}
\label{sec:multi_sync}
\subsection{Dataset details}
\label{sec:dataset_details}
Based on \cite{rstg} we create MultiSyncMNIST. It consists of $10$ frames videos of size $128 \times 128$, where MNIST digits move on a black background. Each video has $5$ moving digits and a subset of them moves synchronously.  Different from the original version, each video could contain multiple instances of the same digit class and any subset can move in the same way. This is done to make it more difficult to distinguish between multiple visual instances. The goal is to detect the smallest and largest digit class among the subset of synchronous digits with each pair of two digits forming a label. In total, we have 55 possible pairs of two digits, and adding a class for videos without synchronous digits results in a $56$-way classification task. For example, if a video contains the digits: $\{2,4,6,7,7\}$ and the subset $\{4,6,7\}$ is moving in the same way, it has the label associated with the pair: $\{4,7\}$. The dataset contains 600k training videos and 10k validation videos.



\subsection{Ablation: Number of Nodes}
\label{sec:ablation_number_of_nodes}

We investigate the effect on the performance of the number of nodes for different environments, of varying difficulty. We conduct experiments varying the complexity of MultiSyncMNIST dataset, by changing the number of moving digit ($D \in \{3,5,9\}$). As expected, we observe (in Table~\ref{tab:mnist-extra-results}) that for good performance, it is necessary to set a number of nodes that exceeds the number of relevant entities in the scene.


\subsection{Ablation: Regions' Size}
\label{sec:ablation_size}

In this subsection, we conduct experiments to investigate the effect of the size of the node regions on the final performance. Each node pools information from latent convolutional features of size $H \times W = 16 \times 16$. We fix the size of each region to $\frac{H}{\lambda}$ where $H = 16$ and $ \lambda \in \{6, 7, 8, 11, 16\}$ and show the results of the corresponding models in Table~\ref{table:sync-size}. Setting $\lambda = 8$ corresponds to regions having approximately the expected values of the regions predicted by the full DyReg-GNN model. We note that the model is relatively robust to reasonable choices of size but the best performance is achieved when the size of each region is dynamically predicted from the input. We also note that by setting $\lambda = H = 16$ we arrive at the standard bilinear interpolation kernel. This setting leads us to a model that is more unstable in training than the others and obtains poorer results. There are two probable reasons for this. First, the regions cover a small area thus they must be more precise to cover small entities while also being unable to cover large entities in their entirely.
Second, the gradients used to update the region parameters are noisier for small regions. This is because, the gradients of the offsets depend on the features of the predicted regions, and for gradients of the offsets to be informative it means that the features in the regions should also be relevant for the final prediction. Smaller regions have a smaller chance of achieving this.

\begin{table*} [t]

\parbox{.48\linewidth}{
\centering
\caption{ 
Results on MultiSyncMNIST when varying the number of nodes on datasets of different complexity (with increasing number of moving digits). It is crucial that the number of nodes exceeds the number of important entities in the scene. In the bottom of the table, we also show two additional baselines with the same number of parameters.}
\begin{tabular}{l l l l}
    \toprule
    Model    & \multicolumn{3}{c@{}}{Dataset} \\
     \# digits(D)   &  D=3 & D=5 & D=9 \\
     \midrule
     DyReg 5 Nodes &  98.2 & 89.6 & 64.4 \\
     DyReg 9 Nodes & 98.1 & 95.1 & 79.3 \\
     DyReg 16 Nodes & \textbf{98.4} & \textbf{95.6} & \textbf{83.0} \\
     
     \midrule
     R18 + Conv-LSTM & - & 89.1 & 50.5 \\
     R18 + NL & - & 93.2 & 67.5\\
\end{tabular}
\label{tab:mnist-extra-results}
}
\hfill
\parbox{.48\linewidth}{


\centering
\caption{Results on Smt-Smt-V2 val. set, using a single $224 \times 224$ central crop. We observe that DyReg-GNN models improve over the TSM backbone and that it is important to have the kick-start given by the distillation to learn multiple dynamic graph modules.
\label{table:smt-distill}
}
\begin{tabular}{l l l l l l}
    \toprule
    Model &  Top 1 & Top 5 \\
    \midrule
    TSM  & 61.1 & 86.5 \\
    DyReg-GNN r3-4-5 &  62.1 & 87.4 \\
    DyReg-GNN r3-4-5 Distill &  62.8 & 87.7 \\
\end{tabular}
}
\end{table*}

\begin{table*}[t]
\parbox{.55\linewidth}{
\centering
\caption{Ablation on MultiSyncMNIST for showing the importance of recurrence for predicting the regions.
\label{tab:ablation_gru}
}
\begin{tabular}{l c}
    \toprule
    Model &  Accuracy \\
    \midrule
    DyReg-GNN without GRU &   91.91\\
    DyReg-GNN &  95.09 \\
 
\end{tabular}
}
\hfill
\parbox{.42\linewidth}{
\centering
\caption{ Comparison to Keypoints-based method on MultiSyncMNIST. 
}
\small
\begin{tabular}{l c c}
    \toprule
    Model & Acc \\
    \midrule
    ResNet-18 & 52.29\\
    Fixed & 78.85 \\
    Keypoints & 90.60 \\
    \midrule
    DyReg-GNN - Pos-Only & 93.41 \\
    DyReg-GNN &  \textbf{95.09}  \\
\end{tabular}
\label{tab:keypoints}
\vspace{-4mm}
}
\end{table*}

\begin{table}[t]
    \centering
    \centering
    \caption{Experiments on MultiSyncMNIST investigating the size of the learned regions. The best performance is obtained when the size is dynamically predicted while the worst is given by a model with the regions kept at the minimum value, corresponding to the standard bilinear interpolation kernel.}
    \begin{tabular}{c c c c c c}
        \toprule
         \theadc{Learnable \\ (Full)} & \theadc{Fix \\ $\lambda=6$}  & \theadc{Fix \\ $\lambda=7$} & \theadc{Fix \\ $\lambda=8$} & \theadc{Fix \\ $\lambda=11$} & \theadc{Fix \\ bilinear}  \\
        \midrule
        95.09 & 93.41  & 94.11 & 94.04 & 94.03 & 90.99 \\
    \end{tabular}
    \label{table:sync-size}
    \centering
\end{table}

\subsection{Ablation: Comparison to Keypoints Extractor}
\label{sec:ablation_keypoints}

We conduct an experiment to compare our dynamic way of generating nodes with a previous method \cite{unsup_obj_cond_img} that detects keypoints from images. 
For a fair comparison, we replace the part in our model that predicts the locations of the node regions with a method similar their encoder.
The rest of our model would remain the same and will be learned in the same way by the video classification loss. We chose the architecture such that the number of parameters remains the same. As in the original paper, this method predicts the positions but keeps the shape fixed, thus we compare it with our module that has fixed shape, although it obtains poorer results than the full model. We report the results in Table~\ref{tab:keypoints}. Note that the dynamic regions obtained using the keypoints method (denoted as Keypoints) improve over the fixed-regions approach, reinforcing the idea that dynamic regions are helpful for relational processing. However, our DyReg-GNN models obtain better results both when the size of the regions is fixed and especially when the size is also predicted. 

\subsection{Ablation: Importance of Recurrence for Region Generation}
We conduct an experiment (Table \ref{tab:ablation_gru}) on the MultiSyncMNIST dataset, where we omit the GRU from Eq. 3, thus predicting the regions at each time step  only from the features of frame. The performance drops from  (DyReg-GNN) to  (DyReg-GNN without GRU). This experiment suggests that the temporal modeling in region generation is important for good performance.

\subsection{Ablation: Ground-Truth Boxes}
\label{sec:ablation_gt_boxes}

To evaluate the quality of our proposed regions on MultiSyncMNIST, we train our model using ground-truth (gt.) boxes instead of generated regions. 
As this task is defined by the exact movements of digits, the gt. boxes represents the ideal regions for the relational model, giving an upper bound for our method.
This oracle model obtains $97.30\%$ accuracy, while the DyReg-GNN model obtains $95.09\%$. Comparing to the other baselines in the main paper, our DyReg-GNN model obtains closer results to the oracle model, proving the utility of the node generation.

\subsection{Comparison to other baselines}
\label{sec:mnist_comp_baselines}

We compare our method to additional baselines by replacing our entire DyReg-GNN module with two other models, as seen in Table~\ref{tab:mnist-extra-results}. 
The first baseline (R18+Conv-LSTM) consists in a convolutional encoder that reduces the spatial dimensions, a shared LSTM applied independently on each spatial position followed by a convolutional decoder. The second baseline (R18-NL) consists in a Non-Local\cite{wang2018non_local} network. Both modules are applied over the same ResNet 18 backbone and have the same number of parameters as DyReg-GNN.
DyReg-GNN surpasses the other baselines and the difference in performance is more significant in the hardest setting.

\begin{table*}[t]
\centering
\caption{Comparison in terms of the number of operations and parameters for a single video of size $224 \times 224$. 
Comparing to assigning nodes to boxes from external detectors (as in I3D+NL+GCN), our module has a smaller computational overhead.
}
\begin{tabular}{l c c c}
    \toprule
    Model & Frames &  FLOPS & Params \\
    \midrule
    I3D~\cite{carreira2017quo} & 32 & 153.0G & 28.0M \\
    I3D+NL~\cite{wang2018non_local} & 32 & 168.0G & 35.3M \\
    I3D+NL+GCN~\cite{wang2018videos_gupta2} & 32 & 303.0G & 62.2M \\
    STM~\cite{jiang2019stm}  & 16 & 66.5G & 24.0M \\
    \midrule
    TSM~\cite{tsm} & 16&  65.8G & 23.9M \\
    TSM + Fixed GNN r4 & 16& 66.3G & 24.9M \\
    TSM + DyReg-GNN r4 & 16 & 66.4G & 25.7M  \\
    TSM + Fixed r3-4-5 & 16 & 67.2G &  26.1M\\
    TSM + DyReg-GNN r3-4-5 & 16 & 67.4G &  28.7M\\
\end{tabular}
\label{table:complexity}
\end{table*}

\section{ Human-Object Interactions}
\label{sec:smt-smt}
\subsection{Distillation for kick-starting the optimisation}
\label{sec:distill}

When training models with multiple DyReg-GNN modules, we observe that only the regions of the last module behaves well, 
thus a single graph module is effectively used. To alleviate this problem we train models with a single graph at different stages, and use their region predictions to distill the larger model, for the first $10\%$ of the training iterations. This kick-starts the learning of all graph modules, improving the overall results, as seen in  Table~\ref{table:smt-distill}.

\subsection{Implementation Details for Detector Experiment}

In the main paper, for the comparison with the regions extracted using object detectors (Section~\ref{para:smt_salient_reg}), we use a Faster RCNN ResNet-50 FPN detector\footnote{ \url{https://github.com/facebookresearch/detectron2}} pre-trained on MSCOCO dataset. 
We extract the top-9 detected boxes based on the confidence score and temporally match them using the hungarian algorithm to maximize the IoU between boxes at consecutive time steps.

\subsection{Object-centric metric}
\label{sec:metrics}

To quantify to what degree the nodes cover existing ground-truth  objects in the scene, we propose the following metric. We measure the distance between the center of the predicted regions and the center of the gt. objects. 
For each node region in each frame, we compute the minimum $L_2$ distance to all gt. object bounding boxes and average all of them.
\begin{equation}
    Dist_p = \frac{1}{N F} \sum_{f=1}^F \sum_{i=1}^N min_j |C_i+\Delta_i - B_j|_2 
\end{equation}
Vice versa we compute for each gt. box the minimum $L_2$ distance to all predicted regions and average all of them.
\begin{equation}
    Dist_r = \frac{1}{N_B F} \sum_{f=1}^F \sum_{j=1}^{N_B} min_i |C_i+\Delta_i - B_j|_2 
\end{equation}
In the previous equations, $F$ is the number of frames in the whole dataset, $N$ the number of nodes, $N_B$ the number of objects in the current frame, $C_i + \Delta_i$ is the center of $i$-th node's region and $B_j$ the center of the $j$-th object in the current frame and we average over the whole dataset.

The first score (representing precision) ensures that all the predicted regions are close to real objects, while the second (recall) ensures that all the objects are close to at least one predicted region. To balance them, we present as our final score their harmonic mean.

 
  



\subsection{Runtime Analysis}
\label{sec:complexity}

    

We compute the number of operations, measured in FLOPS, the parameters and the inference time for our model. We evaluate videos of size $224 \times 224$ in batches of $16$ on a single NVIDIA GTX 1080 Ti GPU.  
TSM backbone, TSM + DyReg-GNN-r4, and DyReg-GNN-r3-4-5 run at $35.7$, $34.8$, and $32.7$ videos per second respectively, showing that our DyReg-GNN module does not add a large overhead over the backbone. 
In Table~\ref{table:complexity}, we compare in terms of number of parameters and operations against other current standard models used in video processing. 
Note that the I3D-based models uses 32 frames but  for our method, the number of operations increases linearly with the number of frames so it is easy to make a fair comparison. The I3D+NL+GCN model counts also the parameters and the operations of the detector module used to extract object boxes. This is characteristic to all the relational models where the nodes are extracted using object detectors. Contrary to this approach, our method has a smaller total complexity by directly predicting salient regions instead of using precise object proposals given by external models.


\fontsize{9.8pt}{10.8pt} \selectfont


\end{document}